\documentclass{article}
\usepackage{ijcai19}

\usepackage{times}
\usepackage{soul}
\usepackage{url}
\usepackage[hidelinks]{hyperref}
\usepackage[small]{caption}
\usepackage{graphicx}
\usepackage{amsmath}
\usepackage{booktabs}
\usepackage{amsfonts}
\usepackage{subfigure}

\title{Two-phase Hair Image Synthesis by Self-Enhancing Generative Model}
\author{
Haonan Qiu$^{12}$\and
Chuan Wang$^3$\and
Hang Zhu$^1$\and
Xiangyu Zhu$^1$\and
Jinjin Gu$^1$\and
Xiaoguang Han$^{12}$\footnote{Corresponding Author}\\
\affiliations
$^1$The Chinese University of Hong Kong, Shenzhen\\
$^2$Shenzhen Research Institute of Big Data, CUHK(SZ) \qquad
$^3$Megvii (Face++) USA\\
\emails\texttt{\{haonanqiu,hangzhu,jinjingu\}@link.cuhk.edu.cn\and uhzoaix@gmail.com\\ wangchuan@megvii.com\and hanxiaoguang@cuhk.edu.cn}
}

\begin{document}
\maketitle
\begin{abstract}
Generating plausible hair image given limited guidance, such as sparse sketches or low-resolution image, has been made possible with the rise of Generative Adversarial Networks (GANs). Traditional image-to-image translation networks can generate recognizable results, but finer textures are usually lost and blur artifacts commonly exist. In this paper, we propose a two-phase generative model for high-quality hair image synthesis. The two-phase pipeline first generates a coarse image by an existing image translation model, then applies a re-generating network with self-enhancing capability to the coarse image. The self-enhancing capability is achieved by a proposed structure extraction layer, which extracts the texture and orientation map from a hair image. Extensive experiments on two tasks, Sketch2Hair and Hair Super-Resolution, demonstrate that our approach is able to synthesize plausible hair image with finer details, and outperforms the state-of-the-art.
\end{abstract}

\section{Introduction}\label{sec:intro}

Accompanied with the success of applying conditional Generative Adversarial Networks (cGANs) \cite{mirza2014conditional} on image to image translation tasks \cite{isola2017image}, generating realistic photos from sparse inputs, such as label maps \cite{chen2017photographic,Lassner:GeneratingPeople:2017} and sketches \cite{Portenier:2018:FDS:3197517.3201393}, nowadays draws much attention of the researchers in both computer graphics and computer vision communities. Portrait image generation, as one of the most popular topics among generative tasks, has been widely studied \cite{wang2018high,karras2017progressive}. Although so, the hair regions, as one of most salient areas, are usually generated with blurry appearances.

\begin{figure}[t]
	\centering
	\subfigure[Hand-drawn sketch]{
		\begin{minipage}[t]{0.47\linewidth}
			{\qquad\includegraphics[height=2.8cm]{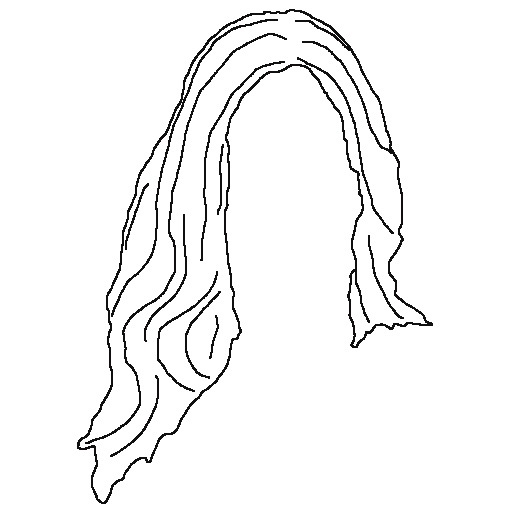}}\label{fig:teaser:a}
		\end{minipage}
	}
	\subfigure[Coarse result by pix2pix]{
		\begin{minipage}[t]{0.47\linewidth}
			{\includegraphics[height=2.8cm]{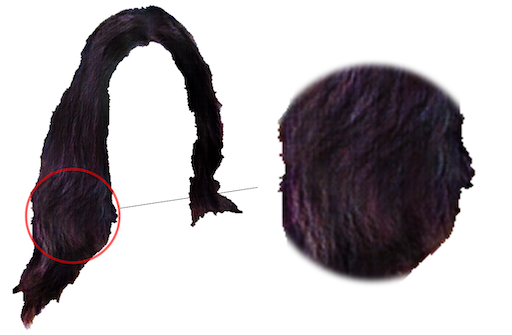}}
		\end{minipage}
	}\\
	\subfigure[Result by pix2pix $+$ style loss]{
		\begin{minipage}[t]{0.47\linewidth}
			{\includegraphics[height=2.8cm]{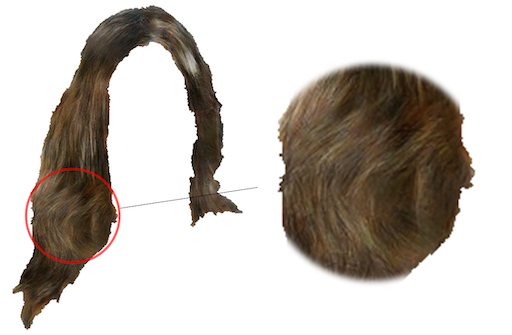}}
		\end{minipage}
	}
	\subfigure[Our result]{
		\begin{minipage}[t]{0.47\linewidth}
			{\includegraphics[height=2.8cm]{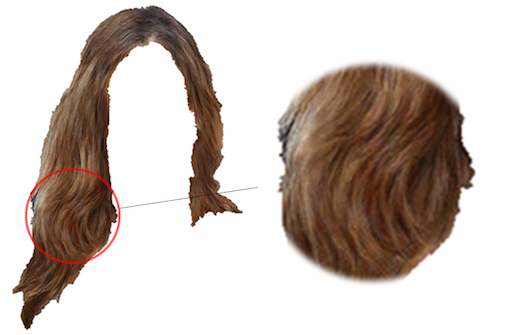}}
		\end{minipage}
	}
	\caption{Given an image with limited guidance, such as a hand-drawn sketch (a), the basic pix2pix framework causes blur artifacts (b). Involving a style loss in training the pix2pix network helps to generate the structure with limited power (c). (d) Results by our two-phase approach, with finer strands and smoother texture synthesized.}
\label{fig:teaser}
\end{figure}
\begin{figure*}[t]
\centering
\includegraphics[width=0.97\linewidth]{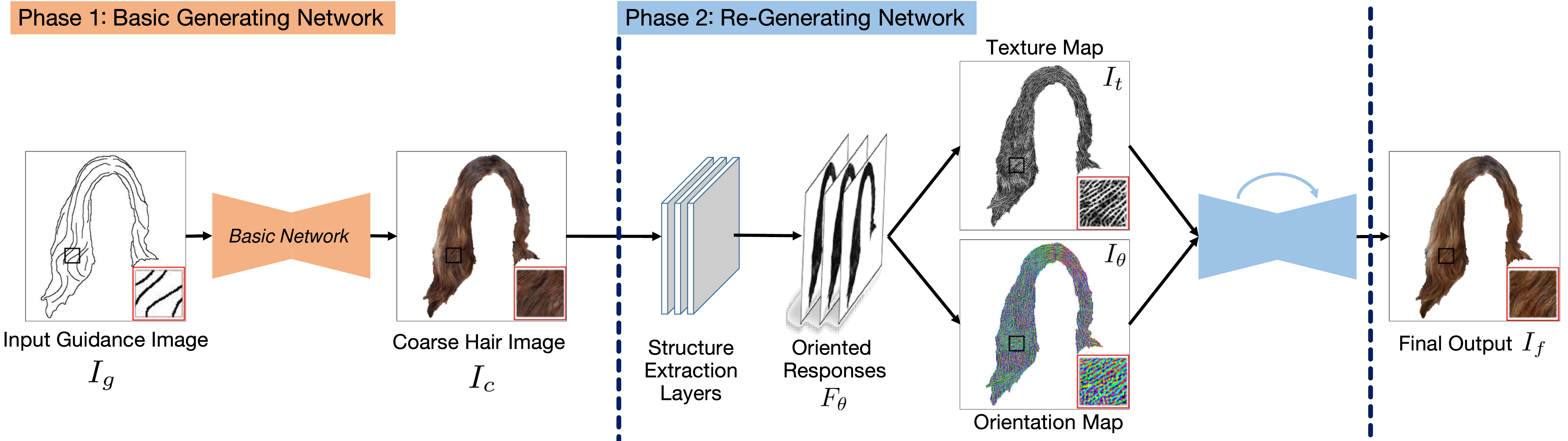}\\
\caption{The architecture of our framework is composed of two phases. In Phase 1, the input image with limited guidance $I_g$ is transformed to a coarse image $I_c$ with a basic net $G_b$. In Phase 2, a re-generating network $G_r$ with self-enhancing capability is attached at the end of $G_b$, and produces the final output $I_f$. Self-enhancement is achieved by our proposed structure extraction layer, extracting texture and orientation map $[I_t, I_\theta]$ as the inputs for a U-net structure. The specific structures of $I_g$, $G_b$ and $G_r$ is task-specific. Here we use the task of Sketch2Hair to illustrate, while it is similar for the task of Hair Super-Resolution.}\label{fig:pipeline} 
\end{figure*}
To touch the heart of this problem, in this paper, we explore approaches to produce realistic hair photographs conditioned on sparse/low-resolution inputs, such as a hair sketch or a downsampled hair image. Hair is very different from other categories of images as its special textures: it is rendered from thousands of long yet thin strands, being full of textured details. This makes the existing cGAN-based methods, such as pix2pix \cite{isola2017image}, fail from two aspects: 1) its discriminator, by encoding the output image into a latent space, guarantees realistic output in a global fashion yet lacks constraints on local sharpness; 2) the input, as the condition of cGANs, is too weak to generate strand-wise pixels.

Based on our experiments, the first issue can be addressed by borrowing feature matching loss from \cite{wang2018high} and style loss from \cite{gatys2016image} to guide the learning procedure. More importantly, to overcome the second challenge and make strands floating out, a self-enhancing generative model is proposed. Our key idea is conducting the generation into two phases: 1) we first utilize the state-of-the-art methods to produce a coarse-level output; 2) the strand-ware structures, such as orientation, is thus extracted. This is then treated as an enhanced condition and fed into a re-generation network, also based on cGANs. To support an end-to-end training of such two-phase network, a novel differentiable texture-extraction layer is embedded in the re-generating network, which enables its capability of self-enhancement.

To validate the effectiveness of the proposed self-enhancing mechanism, two tasks are studied: sketch2hair aims to realistic hair generation from a sparse sketch input; hair super-resolution targets generating high-resolution appearances from a down-sampled image. A high-quality dataset is also built to support conducting these two applications. Both the user study and visual comparisons show the superiorities of the proposed method against all existing ones. Thanks to the proposed structure extraction layer, all of the coarse hair image can be significantly enhanced.

In summary, our main contributions include:
\begin{itemize}
\item A novel self-enhancing module is designed, with which, the hair generation is modeled as an end-to-end two-phase framework. As demonstrated in two tasks, Sketch2Hair and Hair Super-Resolution, this strategy effectively benefits the state-of-the-art generative models for hair synthesis. We foresee this general strategy could be potentially to applied more hair synthesis related tasks.
\item For the task of Sketch2Hair, we are the first application targeting realistic hair image synthesis purely from sketches. This can be regarded as a prototype for real-time sketch-based hairstyle editing.
\item We constructed a high-quality hair dataset including 640 high-resolution hair photos with their corresponding sketches manually drawn. This database will be released upon our paper is accepted to facilitate the research in this field.

\end{itemize}

\section{Related Work}\label{sec:related}

\paragraph{Realistic hair synthesis.} Generating virtual hairstyles is a long-standing research topic in computer graphics due to the important role it plays in representing human characters in games and movies. Most previous works focus on producing 3D hairs, according to user interactions \cite{mao2004sketch,fu2007sketching,hu2015single} or real-captured images \cite{wei2005modeling,jakob2009capturing,chai2016autohair,zhou2018single}.
Given images, thanks to these modeling techniques, the hair can be recovered strand by strand which enables intelligent hair editing \cite{chai2012single,chai2013dynamic} or interpolation \cite{weng2013hair} by performing manipulation in 3D space and then being re-rendered to 2D domain.
Although these methods are able to result in realistic appearances, high computational cost are incurred due to the involvement of 3D matters. To avoid high computational cost in hair rendering, \cite{Wei_2018_ECCV} proposes a deep learning based hair synthesis method, which can generate high-quality results from an edge activation map. However, to obtain the activation map, an input CG hair model is still required for the initial rendering. In comparison, our method involves no 3D rendering module, and relies on a 2D image with sparse information only. With such a limited input, we still synthesize photo-realistic results thanks to the self-enhancing module as proposed.

In contrast, \cite{chen2006generative} put forward a 2D generative sketch model for both hair analysis and synthesis, with which, a hair image can be encoded into a sketch graph, a vector field and a high-frequency band. Taking an image as input, such a multi-level representation provides a straightforward way to synthesize new hairstyles by directly manipulating the sketch. Compared with this work, our approach is feasible to infer a realistic output by only taking in a sparse sketch without any reference photo. We use deep neural networks to achieve spatial consistent conversion from sketches to colorful images instead of traditional image render, which is relatively time-consuming. To our knowledge, we are the first work using cGANs aiming at sketch to hair image synthesis. 

\paragraph{Portrait super-resolution.} Dong et al. \cite{Dong2014Learning,Dong2016Image} pioneered the use of convolutional networks for image super-resolution, achieving superior performance to previous works. Since only three layers of simple convolutional networks are used, SRCNN is still less effective in recovering image details. Later many deeper and more effective networks structures \cite{Kim2016Accurate,Ledig2017Photo,Lim2017Enhanced,Zhang2018Residual,Wang2018ESRGAN} are designed and achieve great success in improving the quality of recovered images. As a hot object, several CNN based architectures \cite{Shizhan2016Deep,Cao2017Attention,Huang2017Wavelet} have been specifically developed for face hallucination. Also vectorization can also be utilized for image super-resolution as in~\cite{lai2009automatic,wang2017video}. To produce more details in results, adversarial learning \cite{Xin2016Ultra,AAAI1714340} is also introduced. Recently, Li et al. \cite{Li2018Learning} propose a semi-parameter approach to reconstruct high quality face image from unknown degraded observation with the help of reference image. However, as an important part of the portrait that fits closely to the face, recovered hair results are always neglected. The hair area recovered by current advanced approaches is always blurred or gelatinous, becoming a short board in portrait Super-Resolution. In this paper, as hair area segmentation is achieved by \cite{Levinshtein2018Real}, we propose an extra hair textured enhancement to improve the recovered hair visual-quality. Our enhanced structure is able to be attached to almost super-resolution methods in an end-to-end manner, which can be regarded as an additional texture enhancement module.

\paragraph{Enhancing technology in generation.} In generative task, spatial consistent conversion between two domains is hard to train especially when the transfer gap is large.
Several enhanced methods from different direction to reduce transfer by dividing the whole conversion into some subtasks.
To stabilize the training process, \cite{chen2017photographic,wang2018high,karras2017progressive} propose the enhanced method that train the network from small scale and then fine tune on large scale.
Also, in semantic segmentation, the strategy of regenerating from preliminary predictions to improve accuracy is widely used \cite{7478072,7410536}.

In low-level vision task, to recover the structure in generated result, \cite{Xu2015Deep,Liu2016Learning} develop networks to approximate a number of filters for edge-preserving. Furthermore, \cite{Pan2018Learning} proposes DualCNN which consists of two parallel branches to recover the structure and details in an end-to-end manner respectively. But in hair generation, generated texture, which can be regarded as structure, is prone to be messed up with some disordered noise. To overcome this problem, we extract structure from the coarse result and use it to regenerate our final result in an end-to-end manner. As far as we know, this enhanced strategy in CNN based generative model has not been proposed before. And our experimental results show this enhanced structure has significant effects to make less blurry area and more meticulous textures.

\section{Network Architecture}\label{sec:network}

Our proposed generative model for hair image synthesis takes an image of limited guidance $I_g$ as input and produces a plausible hair image $I_f$ as output. It enables the high-quality generation by a two-phase scheme. In Phase 1, a basic network $G_b(\cdot)$ as pix2pix in~\cite{isola2017image} is used to generate a coarse result $I_c$, which usually contains little texture and some blur artifacts. Then a re-generating network $G_r(\cdot)$ with self-enhancing capability is applied to produce the high-quality result $I_f$ in Phase 2. We illustrate the entire framework in \figurename~\ref{fig:pipeline}.

\subsection{Basic Generating Network}\label{sec:network:basic}

Given a hair-related image of limited guidance $I_g$, in Phase 1, we conduct an image-to-image translation between $I_g$ and the target domain, i.e. a plausible hair image. The network is commonly a conditional GAN as in~\cite{radford2015unsupervised,isola2017image}. Specifically, in the tasks of Sketch2Hair (S2H) and Hair Super-Resolution (HSR), $I_g$ are a sparse sketch image (\figurename~\ref{fig:teaser:a}) and low-resolution image (\figurename~\ref{fig:SRGAN:input}) respectively. Accordingly, the target image $I_c$ is of the same resolution as $I_g$ for S2H, and of a user-specified higher resolution for HSR. The network structures are also task specified. For example, in HSR the structure contains several more upsampling layers compared with S2H. We refer readers to~\cite{isola2017image} for detailed descriptions of image-to-image translation networks for the two tasks. Also, for simplicity, we use "basic net" or "basic network" for an abbreviations of "basic generating network" in this paper.

\subsection{Self-Enhancing Re-Generating Network}\label{sec:network:self}
The hair image produced by Phase 1, $I_c$, is usually recognisable with core structure and close to the target. However, it is still far from plausible due to its lack of gloss, texture and strand-like style. To generate a high-quality hair image $I_f$, we further feed $I_c$ into a re-generating network $G_r(\cdot)$ with self-enhancing capability, which is achieved by a newly-introduced \textit{structure extraction layer} as follows.

\begin{figure}[!t]
	\centering
	\subfigure[$\theta = 0$]{
		\begin{minipage}[t]{0.22\linewidth}
			{\includegraphics[height=2cm]{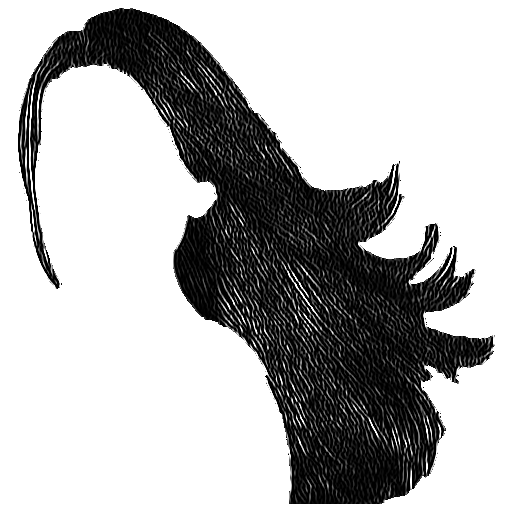}}
		\end{minipage}
	}
	\subfigure[$\theta = \frac{\pi}{4}$]{
		\begin{minipage}[t]{0.22\linewidth}
			{\includegraphics[height=2cm]{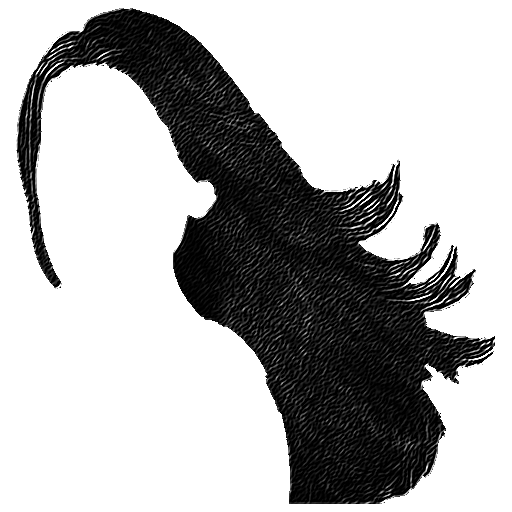}}
		\end{minipage}
	}
	\subfigure[$\theta = \frac{\pi}{2}$]{
		\begin{minipage}[t]{0.22\linewidth}
			{\includegraphics[height=2cm]{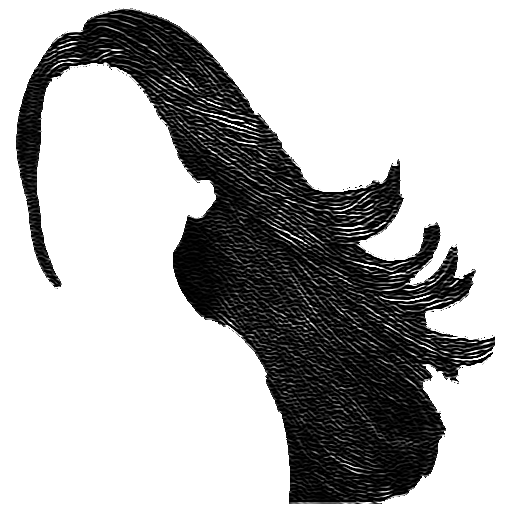}}
		\end{minipage}
	}
	\subfigure[$\theta = \frac{3\pi}{4}$]{
		\begin{minipage}[t]{0.22\linewidth}
			{\includegraphics[height=2cm]{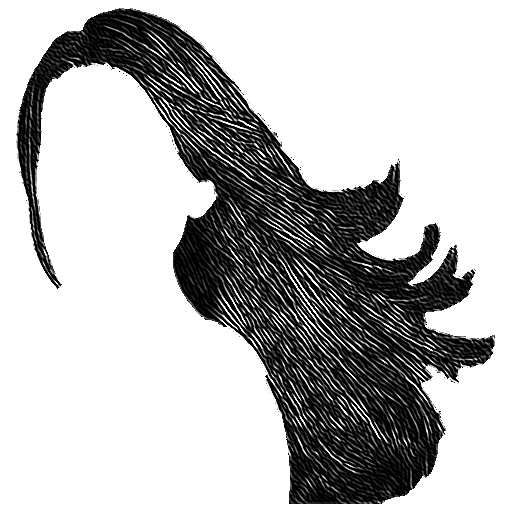}}
		\end{minipage}
	}
	\caption{Gabor responses in different orientations. Hairline has a large response for its corresponding oriented Gabor filter. With this property, the core structure of hair can be extracted by the designed Gabor filter bank.} 
\label{fig:gabor}
\end{figure}

\subsubsection{Structure Extraction Layer}
Given a hair image $I$, we follow the pipeline in~\cite{chai2012single} to filter $I$ by a set of oriented filters $\{K_\theta\}$, generating a set of oriented responses $\{F_\theta\}$, i.e.
\begin{equation}
F_\theta(i,j) = (K_\theta * I)(i,j)
\end{equation}
where $(i,j)$ is the pixel location, and $\theta$ is the orientation angle. For each pixel location $(i,j)$, we pick up the $\theta$ which maximizes $F_\theta(i,j)$, and the corresponding maximal value to obtain an orientation map $I_{\theta}(i,j)$ and a texture map $I_t(i,j)$, respectively, i.e.
\begin{align}
I_{\theta}(i,j) &= \arg\max_\theta \{ F_\theta(i,j) \} \\
I_t(i,j) &= \max_\theta \{ F_\theta(i,j) \}
\end{align}

\begin{figure}[!t]
	\centering
	\subfigure[$I$]{
		\begin{minipage}[t]{0.3\linewidth}
			{\includegraphics[height=2.8cm]{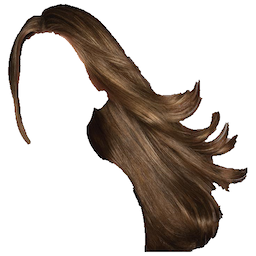}}
		\end{minipage}
	}
	\subfigure[$I_t$]{
		\begin{minipage}[t]{0.3\linewidth}
			{\includegraphics[height=2.8cm]{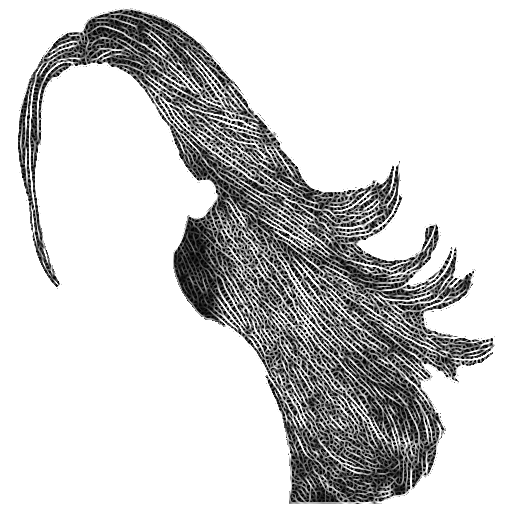}}
		\end{minipage}
	}
	\subfigure[$I_\theta$]{
		\begin{minipage}[t]{0.3\linewidth}
			{\includegraphics[height=2.8cm]{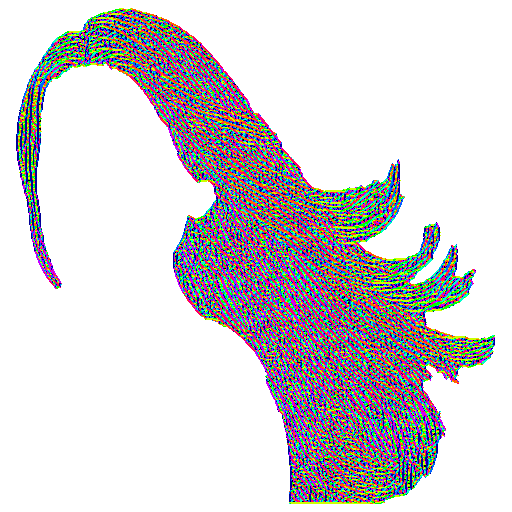}}
		\end{minipage}
	}
	\caption{Visualization of texture map and orientation map. (a) Input image. (b)(c) Texture map and orientation map of (a) respectively. Here orientation map is colorized for visualization. }
\label{fig:orientation}
\end{figure}

In our paper, we utilize 8 even-symmetric cosine Gabor kernels as the filter bank $\{K_\theta\}$, where the orientation angle $\theta$ being evenly sampled between $0^\circ$ and $180^\circ$. Specifically, the cosine Gabor kernel at angle $\theta$ is defined as:
\begin{equation}
K_{\theta}(u ,v)=\exp\left ( -\frac{1}{2}\left [ \frac{\tilde{u}^{2}}{\sigma _{u}^{2}}+\frac{\tilde{v }^{2}}{\sigma_{v }^{2}} \right ] \right ) \cdot \cos\left ( \frac{2\pi \tilde{u}}{\lambda } \right )
\end{equation}
where $\tilde{u}=u\cos\theta+v\sin\theta$ and $\tilde{v}=-u\sin\theta +v\cos\theta$. $\sigma_u$, $\sigma_v$, $\lambda$ are hyper-parameters, and we simply set them 1.8, 2.4, 4 in all of our experiments respectively. We denote the operations stated above as $[I_\theta, I_t] \leftarrow \mathbf{g}(I)$. \figurename~\ref{fig:gabor} illustrates Gabor responses in 4 orientations, and \figurename~\ref{fig:orientation} visualizes an example of texture map and orientation map.

In practice, we found that $I_t$ initially extracted from $I_c$ contains most textures but some blurry artifacts in $I_c$ also retains. To tackle this issue, we duplicate the operation $\mathbf{g}$, first applying it to $I_c$ to produce $I^0_t$ and then re-applying it to $I^0_t$ to obtain the final $I_t$ and $I_\theta$, i.e.
\begin{equation}
[I_\theta, I_t] \leftarrow \mathbf{g}(I^0_t), \quad \text{where}\ \ [I^0_\theta, I^0_t] \leftarrow \mathbf{g}(I_c) \label{eq:operation:g}
\end{equation}

\begin{figure}[!t]
	\centering
	\subfigure[$I_g$]{
		\begin{minipage}[t]{0.47\linewidth}
			{\qquad\includegraphics[height=2.8cm]{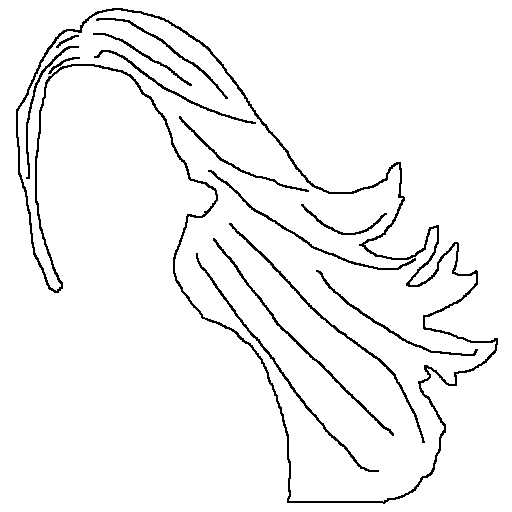}}
		\end{minipage}
	}
	\subfigure[$I_c$]{
		\begin{minipage}[t]{0.47\linewidth}
			{\includegraphics[height=2.8cm]{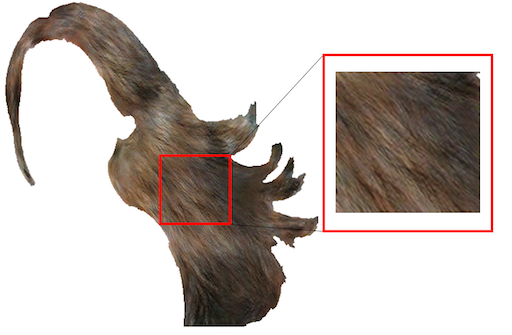}}
		\end{minipage}
	}\\
	\subfigure[$I^0_f$]{
		\begin{minipage}[t]{0.47\linewidth}
			{\includegraphics[height=2.8cm]{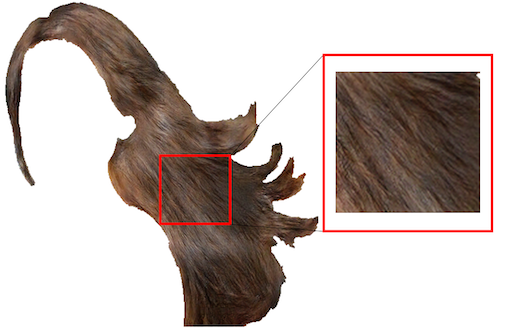}}
		\end{minipage}
	}
	\subfigure[$I_f$]{
		\begin{minipage}[t]{0.47\linewidth}
			{\includegraphics[height=2.8cm]{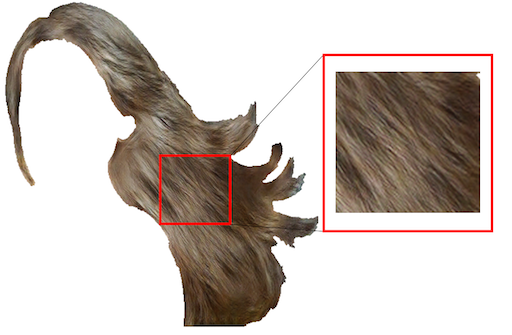}}
		\end{minipage}
	}
\caption{Comparison of the final results, by using structure information $[I^0_t,I^0_\theta]$ and $[I_t,I_\theta]$. (a) An input sketch image. (b) The coarse result generated by basic network $G_b$. (c)(d) The final results generated by using the extracted structure information $[I^0_t,I^0_\theta]$ and $[I_t,I_\theta]$, respectively. Compared with (c), (d) is in higher quality in terms of textures and gloss.}
\label{fig:duplicate:func}
\end{figure}

We demonstrate the effectiveness of using $[I_t,I_\theta]$ instead of $[I^0_t,I^0_\theta]$ in \figurename~\ref{fig:duplicate:func}.

\subsubsection{Network Structure and Optimization Goal}
$[I_t, I_\theta]$ obtained by the structure extraction layer now contains not only the structure but also the detailed information of hair. To take fully advantage of these high-level and low-level features, we apply U-Net architecture~\cite{ronneberger2015u-net:} which is composed of fully convolutional neural networks and skip connections, and produce $I_f = G_r(I_c)$. Also, the optimization goal is formulated as a combination of several types of losses, including pixel loss $L_{pixel}$, adversarial loss $L_{adv}$, style loss $L_{style}$, and feature matching loss $L_{FM}$.

\paragraph{Pixel-level reconstruction.} The output produced by our re-generating network needs to respect the ground truth in pixel-level as the traditional methods do~\cite{radford2015unsupervised,isola2017image}. This is achieved by computing the $L_1$ loss between $I_f$ and ground truth image $I_{gt}$ as in~\cite{wang2018video,meng2018mganet,wang2019gif2video}. Similarly, to encourage sharp and varied hair generation, we also introduce adversarial loss $L_{adv}$ as in~\cite{goodfellow2014generative}. These two losses are written as
\begin{align}
L_{pixel} &= \frac{1}{H\times W\times C}\| I_{gt} - G_r(I_c) \|_1 \\
L_{adv} &= -\sum_{I_c} \log D(G_r(I_c))
\end{align}
where $H,W,C$ are the sizes of the image, and $D$ is a discriminator of GAN. 

\paragraph{Style reconstruction.}
One core issue of synthesizing plausible hair image is generating silky and glossy hair style. Training with $L_{pixel}$ and $L_{adv}$ forces the network paying too much attention to the per-pixel color transfer and limited high-frequency generation, instead of reconstructing realistic styles of hair. Therefore, we incorporate the associated constraint by measuring the input-output distance in a proper feature space, which is sensitive to the style change such as integral colors, textures and common patterns, while relatively robust to other variations. Specifically, we define $L_{style}$ the following manner.

Denote $\psi$ as a pre-trained network for feature extraction and $\psi^i(x)$ is the feature map at its $i_{th}$ layer, whose shape is $C_i \times H_i \times W_i$. A Gram matrix of shape $C_i \times C_i$ as introduced in~\cite{gatys2016image} is built with its entry defined as:
\begin{equation}
	\mathbf{G}^{i}(x)_{j,k}=\frac{1}{C_{i}H_{i}W_{i}}\sum_{h= 1}^{H_{i}}\sum_{w = 1}^{W_{i}}\psi^i(x)_{j,h,w}\cdot\psi^i(x)_{k,h,w}
\end{equation}
Then the style reconstruction loss at the $i_{th}$ layer is formulated as the squared Frobenius norm of the difference between the Gram matrices of the output and target image:
\begin{equation}
L^i_{style}(I_c, I_{gt}) = \left \| \mathbf{G}^i(G_r(I_c)) - \mathbf{G}^i(I_{gt}) \right \|^2_F
\end{equation}
In practice, we apply a pre-trained VGG-16 network as $\psi$, and accumulate the style losses on its two layers (relu2\_2, relu3\_3) to formulate $L_{style}$. In addition, for efficiency, we reshape $\psi^i(x)$ into a matrix $\chi$ of shape $C_{i}\times H_{i}W_{i}$, so as to obtain $\mathbf{G}^i(x)=\chi \chi ^{T}/ C_{i} H_{i}W_{i}$.

\begin{figure}[t]
  \centering
  \subfigure[$I_t$ without $L_{texture}$]{
  \includegraphics[width=0.48\linewidth]{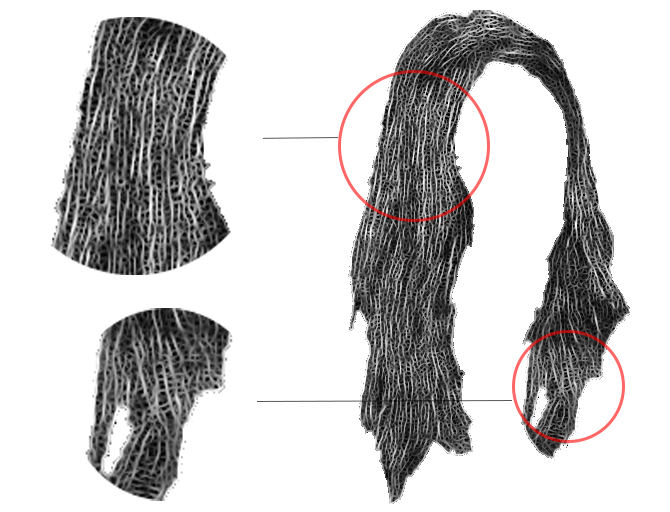}}
  \subfigure[$I_t$ with $L_{texture}$]{
  \includegraphics[width=0.48\linewidth]{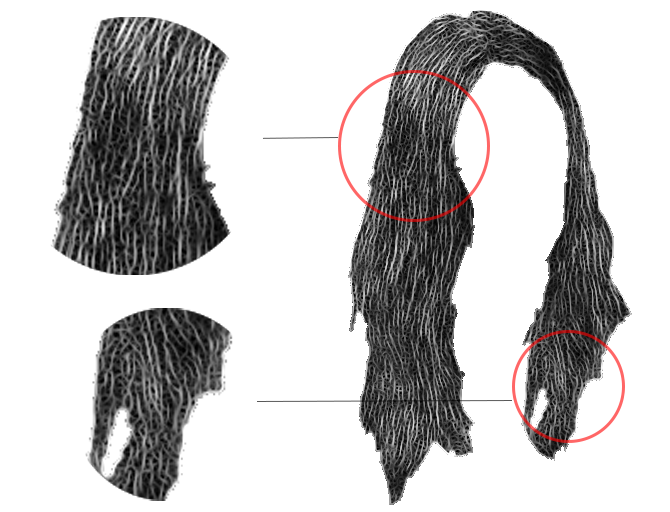}}
  \caption{Texture map extracted without vs. with $L_{texture}$. Compared with (a), clearer hairlines are generated in (b) with the guide of texture loss.}\label{fig:texture-map-comp}
\end{figure}

\paragraph{Feature matching.} As discussed in \cite{wang2018high}, high-resolution image synthesis poses a great challenge to the GAN discriminator design. To differentiate high-resolution real and synthesized images, the discriminator $D$ needs to be deeper to have a large receptive field. Thus it is difficult for $D$ to penalize the difference in detail.
A featuring matching loss \cite{wang2018high} defined on the feature layers extracted by $D$ can well simplify this problem, i.e.

\begin{align}
L_{FM}(G_r, D) = \mathbb{E}_{I_c}\sum_{i=1}^T{\frac{1}{N_i}\left \| D^i(I_{gt}) - D^i(G_r(I_c)) \right \|_{1}}
\end{align}
where $D^i$ is the feature map in the $i_{th}$ layer of $D$, $N_i$ is the number of elements in each layer and $T$ is the total number of layers.

Our full objective combines all of the losses above as
\begin{align}
\min_{G_r} \left( w_{1}L_{pixel}+w_{2}L_{adv}+w_{3}L_{style}+w_{4}L_{FM}\right) \label{eq:objective}
\end{align}
where $w_1 \sim w_4$ controls the importance of the 4 terms.

\subsection{Training Strategy}
The basic network and the re-generating network are trained separately at the very beginning. For the re-generating network, we first directly extract the structure information $[I_t, I_\theta]$ from the ground truth $I_{gt}$. When recognizable coarse images are available from the trained basic network, we replace the source of $[I_t, I_\theta]$ from ground truth image to the coarse image $I_c$. This strategy enables the trained re-generating network to see enough data to avoid easily over-fitting. Finally, we connect the two networks and fine-tune them jointly with a reduced learning rate.

\begin{figure}
	\centering
	\subfigure[]{
		\begin{minipage}[t]{0.22\linewidth}
			{\includegraphics[height=2cm]{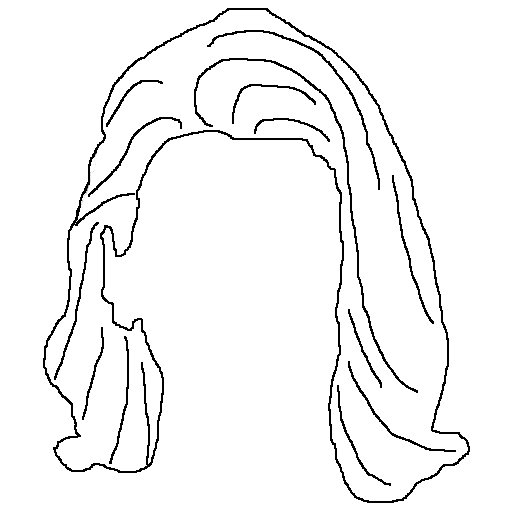}}
		\end{minipage}
	}
	\subfigure[]{
		\begin{minipage}[t]{0.22\linewidth}
			{\includegraphics[height=2cm]{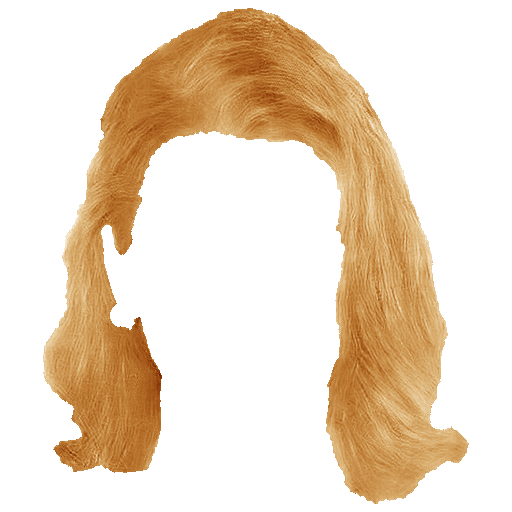}}
		\end{minipage}
	}
	\subfigure[]{
		\begin{minipage}[t]{0.22\linewidth}
			{\includegraphics[height=2cm]{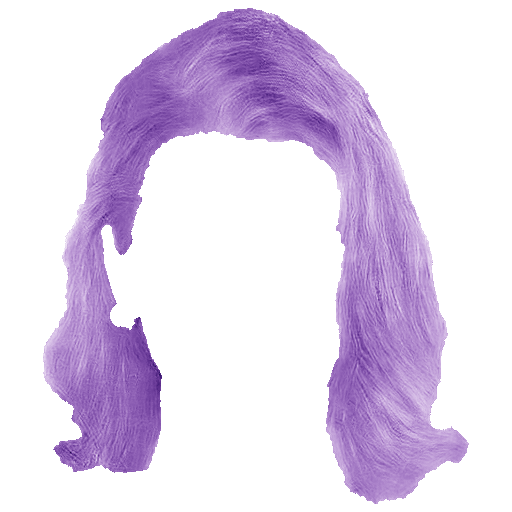}}
		\end{minipage}
	}
	\subfigure[]{
		\begin{minipage}[t]{0.22\linewidth}
			{\includegraphics[height=2cm]{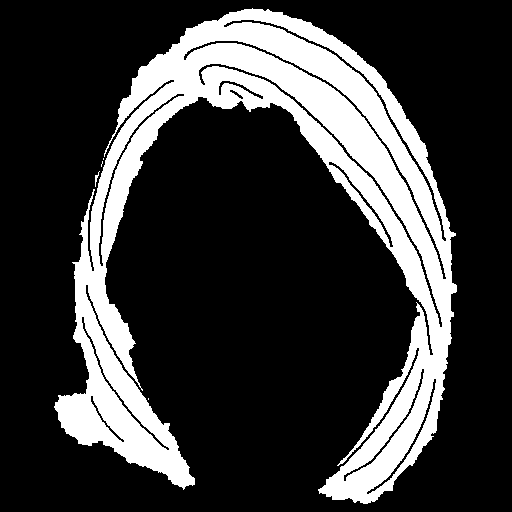}}
		\end{minipage}
	}\\
    \subfigure[]{
		\begin{minipage}[t]{0.22\linewidth}
			{\includegraphics[height=2cm]{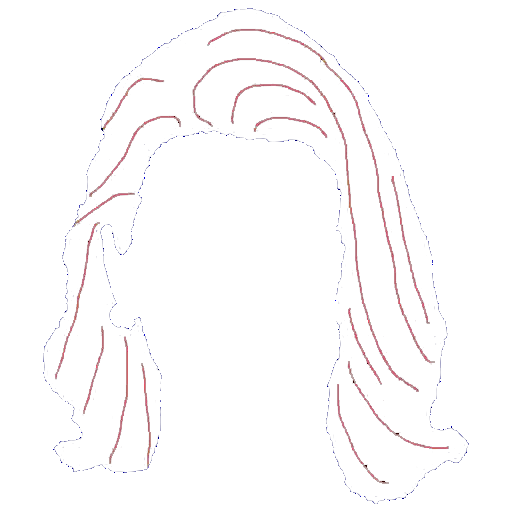}}
		\end{minipage}
	}
	\subfigure[]{
		\begin{minipage}[t]{0.22\linewidth}
			{\includegraphics[height=2cm]{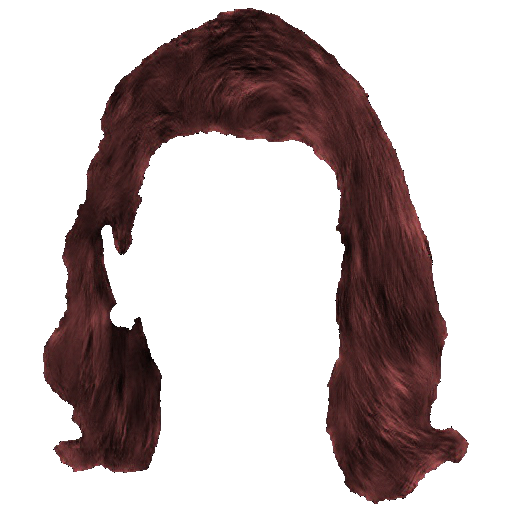}}
		\end{minipage}
	}
	\subfigure[]{
		\begin{minipage}[t]{0.22\linewidth}
			{\includegraphics[height=2cm]{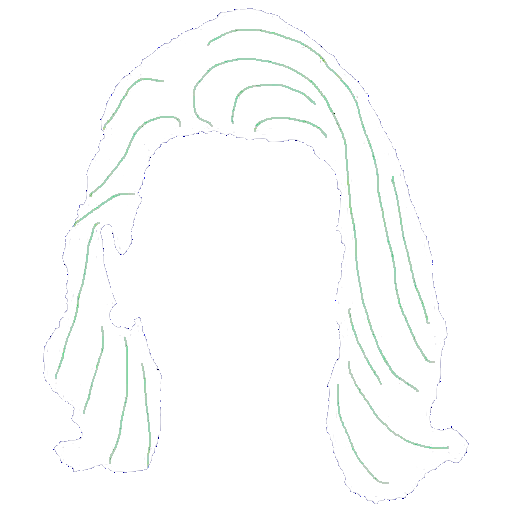}}
		\end{minipage}
	}
	\subfigure[]{
		\begin{minipage}[t]{0.22\linewidth}
			{\includegraphics[height=2cm]{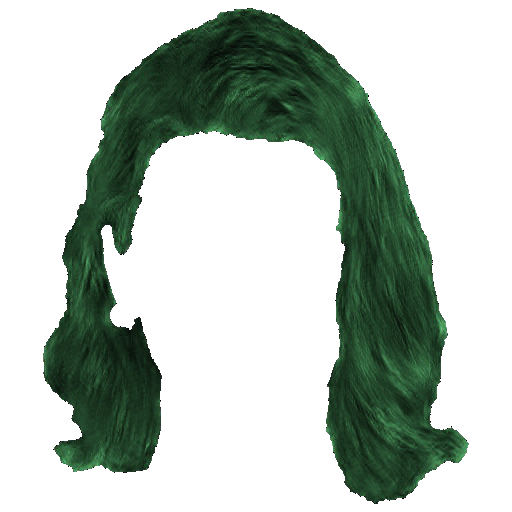}}
		\end{minipage}
	}
	\caption{Colorized results by two solutions, (a) to (d) by postprocessing, i.e. tuning tones in Photoshop, and (e) to (h) by data-driven approach. (e) and (g) are the input sketch images with color strokes, (f) and (h) are their corresponding outputs.}
\label{fig:color:hair}
\end{figure}

\section{Applications}\label{sec:applications}

\subsection{Sketch2Hair}
The input for the Sketch2Hair task is a binary image $I_g$ with 2 channels. Its first channel is a mask $M \in \{0,1\}$ where 0 and 1 fill the outside and inside of the contour of the hair, respectively. Its second channel $S \in \{0,1\}$ encodes the strokes $(0)$ indicating
the dominant growth directions of strands. As aforementioned, in Phase 1, $I_g$ is fed into a basic generating network $G_b$ to produce a coarse result $I_c$, and then a re-generating network $G_r$ is applied to produce the final result $I_f$.

For $G_b$, we choose a deep U-net \cite{ronneberger2015u-net:} that convolves the input sketch into $1\times1$ feature maps and then deconvolutes it iteratively to generate $I_c$ of the same size as $I_g$. The reason behind is that, we require the network has enough capability to imagine the whole hairlines, given the rough flow information in the sparse strokes. It should not trivially view the strokes as the only hairlines for the synthesized image, in which case, the network may only render some background color between them as in~\cite{chen2017photographic}. Deconvolution from $1\times1$ feature maps makes each recovered pixel sees all input pixels, and deep structure makes the network has more parameters to learn from the data.

Additionally, during training $G_b$, we add a texture loss $L_{texture}$ into the existing $L_{pixel}$ and $L_{adv}$, to enforce $I_c$ containing as much as texture information as possible. Specifically, texture loss penalizes the texture difference between $I_{gt}$ and $I_c$, and is written as
\begin{align}
L_{texture}=\left \| I_{t}(I_{gt})-I_{t}(I_{c}) \right \|_{1}
\end{align}
where $I_t(x)$ means the texture map extracted from $x$ as defined similarly in Eq.~\eqref{eq:operation:g}. We illustrate two texture maps extracted with vs. without $L_{texture}$ after optimizing $G_b$ in \figurename~\ref{fig:texture-map-comp}. As seen, with this loss included, clearer hairlines are present in the texture map, facilitating the high-quality image generation in the successive network.

Then we attach $G_r$ to the end of $G_b$ to improve the final results. We list extensive comparisons in Section~\ref{sec:exps} to demonstrate the effectiveness of $G_r$, specifically in \figurename~\ref{fig:teaser} and~\ref{fig:comparisons}.

$I_f$ generated by $G_r$ is usually drab and lacks of rich colors, while a color-specified synthesized hair image is often required. To this end, we provide two solutions for it. One is to tune the tone of $I_f$ in image processing software like Photoshop, which preserves the gloss and silky style as in $I_f$. The other one is data-driven similar to~\cite{Sangkloy2017Scribbler}, i.e. augment the training pairs into colorized version, so as to enable the networks to learn the color correspondence. Experimental results in \figurename~\ref{fig:color:hair} show that the two solutions achieve pleasing outputs.

\subsection{Hair Super-Resolution}
Super-resolution for human portrait images is in higher demand due to the increasing use of digital camera apps, where hair super-resolution plays a key role. In this task, the input image $I_g$ is a low-resolution image and $I_f$ is in higher resolution. Considering visual performance, we choose SRGAN~\cite{Ledig2017Photo} and ESRGAN~\cite{Wang2018ESRGAN} as our basic nets to produce $I_c$. In experiments, we found that if the input image is too small, the basic nets will fail in producing accurate results, obtaining blur artifacts. With our re-generating network introduced, its self-enhancing capability enables the generation of finer textures and fewer artifacts exist in $I_f$. Note that here, we also feed a bicubic upsampled version $\tilde{I}_g$ of input $I_g$ to $G_r$, because $[I_t, I_\theta]$ contains only structure instead of color information.

We also list the comparison results in \figurename~\ref{fig:SRGAN} to demonstrate the effectiveness of our approach.

\begin{figure}
	\centering
	\subfigure{
		\begin{minipage}[t]{0.09\linewidth}
			{\includegraphics[width=0.9cm]{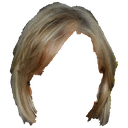}}
			{\includegraphics[width=0.9cm]{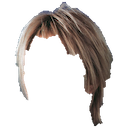}}
			{\includegraphics[width=0.9cm]{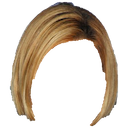}}
			{\includegraphics[width=0.9cm]{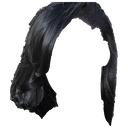}}
		\end{minipage}
	}
	\subfigure{
		\begin{minipage}[t]{0.09\linewidth}
			{\includegraphics[width=0.9cm]{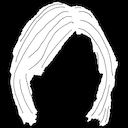}}
			{\includegraphics[width=0.9cm]{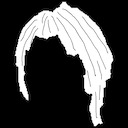}}
			{\includegraphics[width=0.9cm]{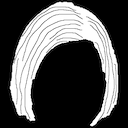}}
			{\includegraphics[width=0.9cm]{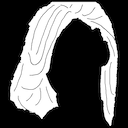}}
		\end{minipage}
	}
	\subfigure{
		\begin{minipage}[t]{0.09\linewidth}
			{\includegraphics[width=0.9cm]{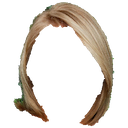}}
			{\includegraphics[width=0.9cm]{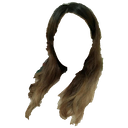}}
			{\includegraphics[width=0.9cm]{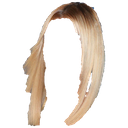}}
			{\includegraphics[width=0.9cm]{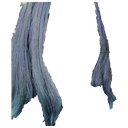}}
		\end{minipage}
	}
	\subfigure{
		\begin{minipage}[t]{0.09\linewidth}
			{\includegraphics[width=0.9cm]{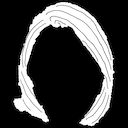}}
			{\includegraphics[width=0.9cm]{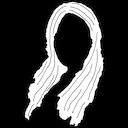}}
			{\includegraphics[width=0.9cm]{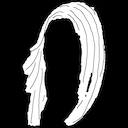}}
			{\includegraphics[width=0.9cm]{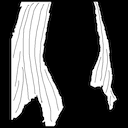}}
		\end{minipage}
	}
	\subfigure{
		\begin{minipage}[t]{0.09\linewidth}
			{\includegraphics[width=0.9cm]{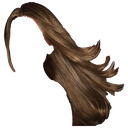}}
			{\includegraphics[width=0.9cm]{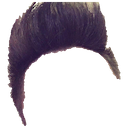}}
			{\includegraphics[width=0.9cm]{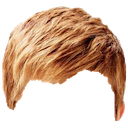}}
			{\includegraphics[width=0.9cm]{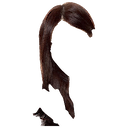}}
		\end{minipage}
	}
	\subfigure{
		\begin{minipage}[t]{0.09\linewidth}
			{\includegraphics[width=0.9cm]{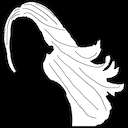}}
			{\includegraphics[width=0.9cm]{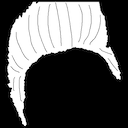}}
			{\includegraphics[width=0.9cm]{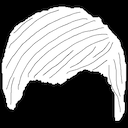}}
			{\includegraphics[width=0.9cm]{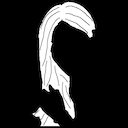}}
		\end{minipage}
	}
	\subfigure{
		\begin{minipage}[t]{0.09\linewidth}
			{\includegraphics[width=0.9cm]{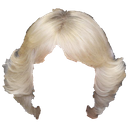}}
			{\includegraphics[width=0.9cm]{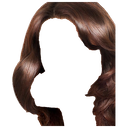}}
			{\includegraphics[width=0.9cm]{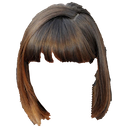}}
			{\includegraphics[width=0.9cm]{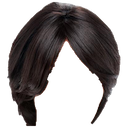}}
		\end{minipage}
	}
	\subfigure{
		\begin{minipage}[t]{0.09\linewidth}
			{\includegraphics[width=0.9cm]{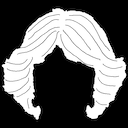}}
			{\includegraphics[width=0.9cm]{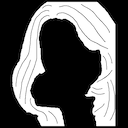}}
			{\includegraphics[width=0.9cm]{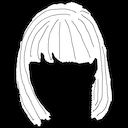}}
			{\includegraphics[width=0.9cm]{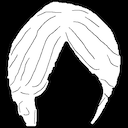}}
		\end{minipage}
	}
	\caption{A glance of the hair dataset, including HD hair images (odd columns) and the corresponding sketches (even columns).}
\label{fig:dataset}
\end{figure}

\section{Experimental Results}\label{sec:exps}

\begin{figure*}
	\centering
	\subfigure[input]{
		\begin{minipage}[t]{0.12\linewidth}
			{\includegraphics[width=2.3cm]{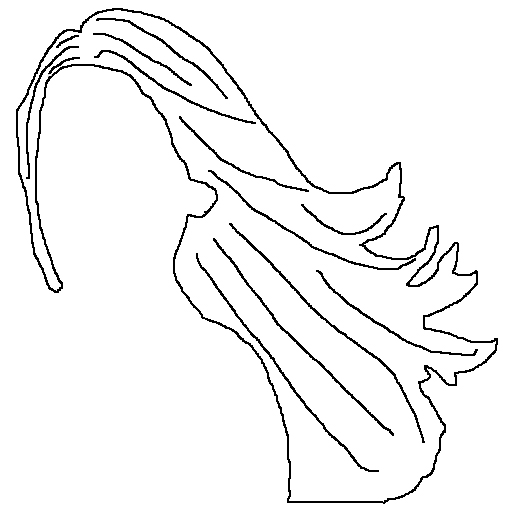}}
			{\includegraphics[width=2.3cm]{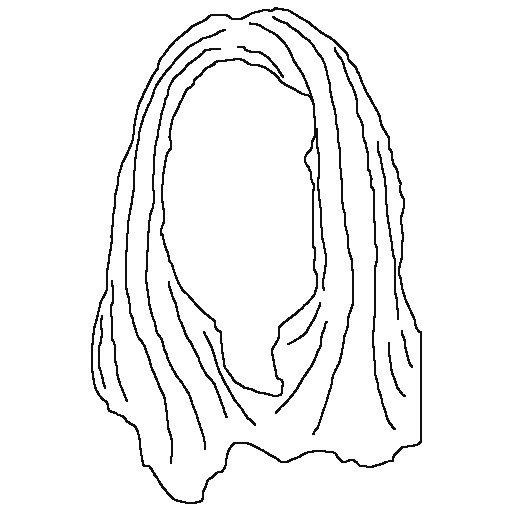}}
			{\includegraphics[width=2.3cm]{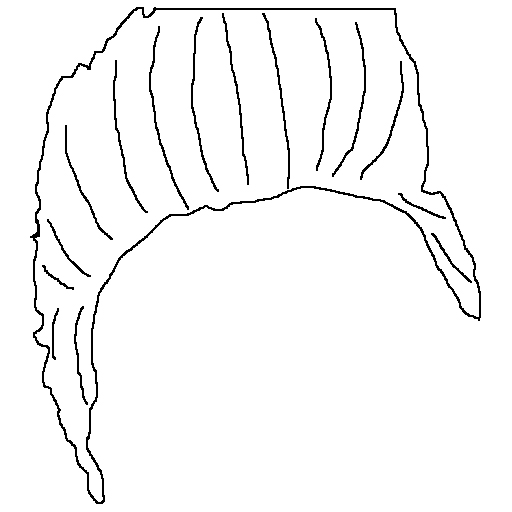}}
			{\includegraphics[width=2.3cm]{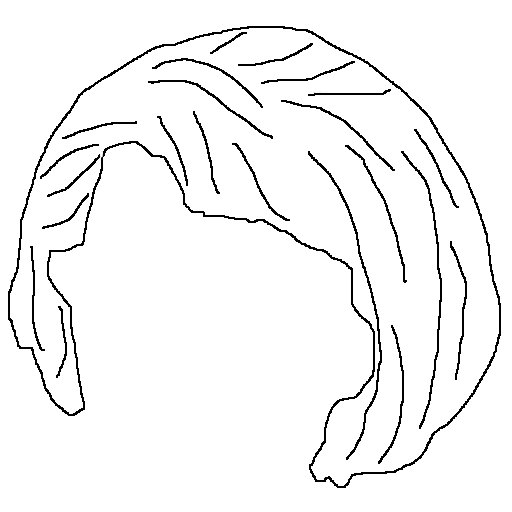}}
		\end{minipage}
	}
	\subfigure[pix2pix]{
		\begin{minipage}[t]{0.12\linewidth}
			{\includegraphics[width=2.3cm]{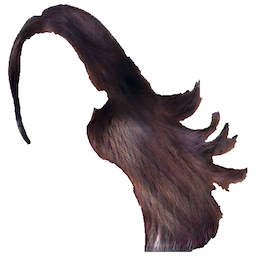}}
			{\includegraphics[width=2.3cm]{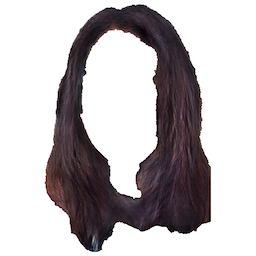}}
			{\includegraphics[width=2.3cm]{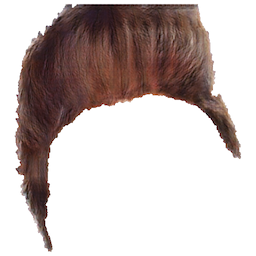}}
			{\includegraphics[width=2.3cm]{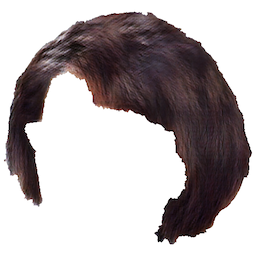}}
		\end{minipage}
	}
	\subfigure[CRN]{
		\begin{minipage}[t]{0.12\linewidth}
			{\includegraphics[width=2.3cm]{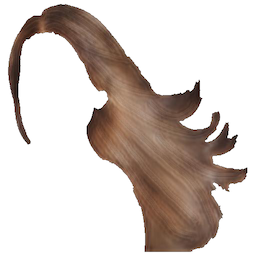}}
			{\includegraphics[width=2.3cm]{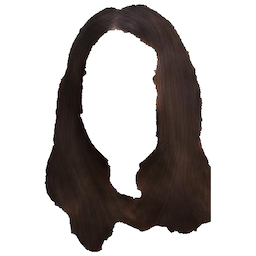}}
			{\includegraphics[width=2.3cm]{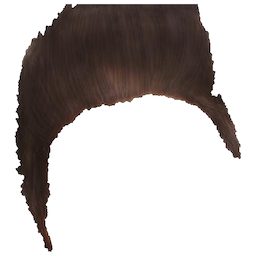}}
			{\includegraphics[width=2.3cm]{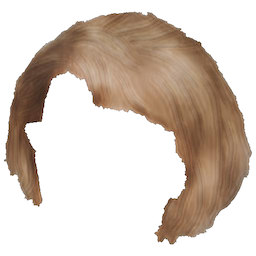}}
		\end{minipage}
	}
	\subfigure[pix2pixHD]{
		\begin{minipage}[t]{0.12\linewidth}
			{\includegraphics[width=2.3cm]{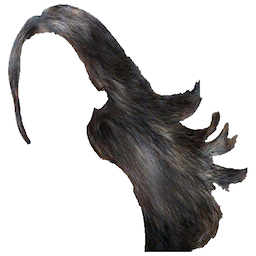}}
			{\includegraphics[width=2.3cm]{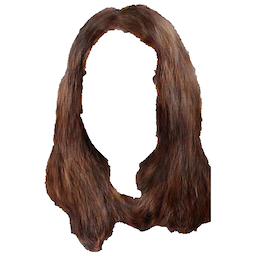}}
			{\includegraphics[width=2.3cm]{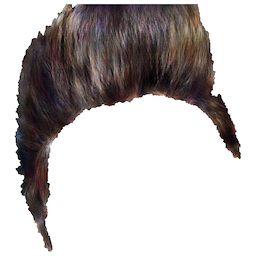}}
			{\includegraphics[width=2.3cm]{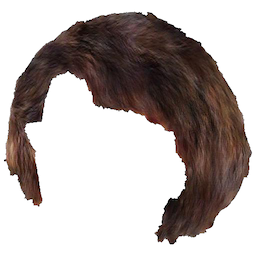}}
		\end{minipage}
	}
	\subfigure[pix2pixStyle]{
		\begin{minipage}[t]{0.12\linewidth}
			{\includegraphics[width=2.3cm]{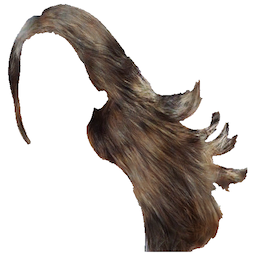}}
			{\includegraphics[width=2.3cm]{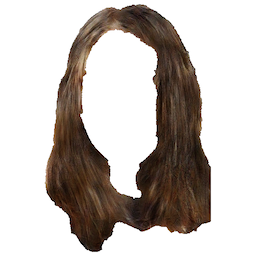}}
			{\includegraphics[width=2.3cm]{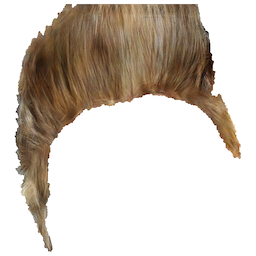}}
			{\includegraphics[width=2.3cm]{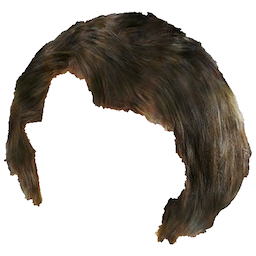}}
		\end{minipage}
	}
	\subfigure[ours]{
		\begin{minipage}[t]{0.12\linewidth}
			{\includegraphics[width=2.3cm]{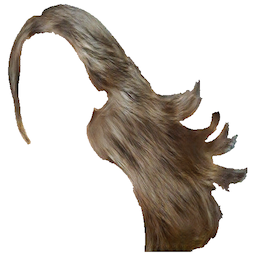}}
			{\includegraphics[width=2.3cm]{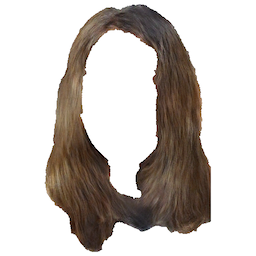}}
			{\includegraphics[width=2.3cm]{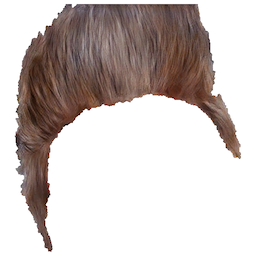}}
			{\includegraphics[width=2.3cm]{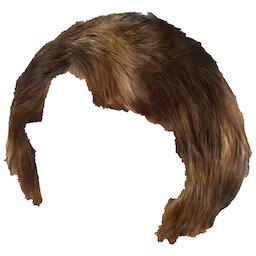}}
		\end{minipage}
	}
	\subfigure[ground truth]{
		\begin{minipage}[t]{0.12\linewidth}
			{\includegraphics[width=2.3cm]{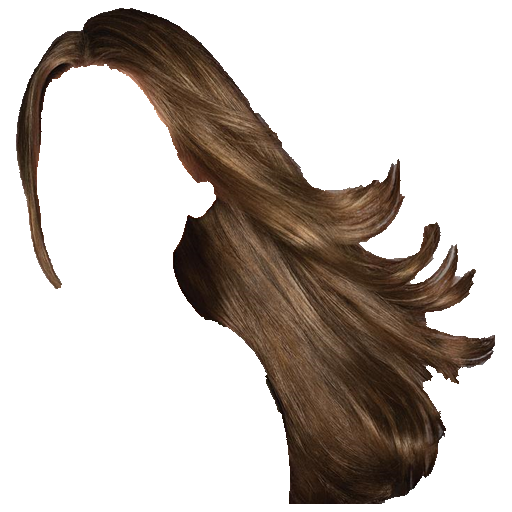}}
			{\includegraphics[width=2.3cm]{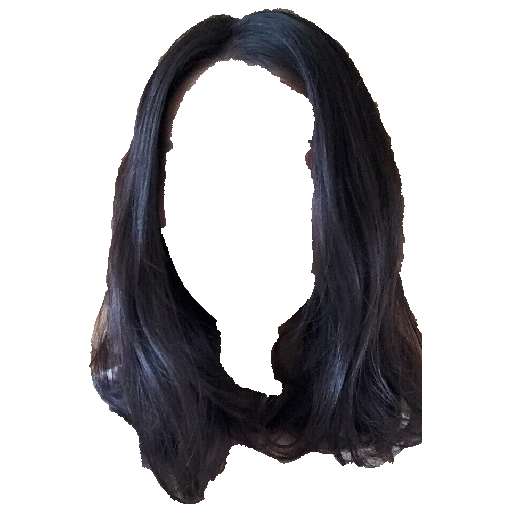}}
			{\includegraphics[width=2.3cm]{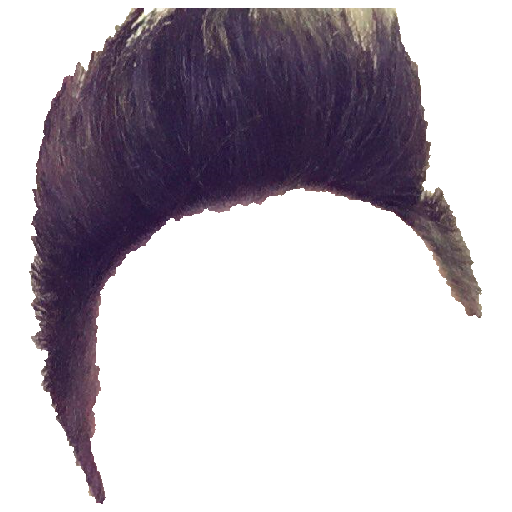}}
			{\includegraphics[width=2.3cm]{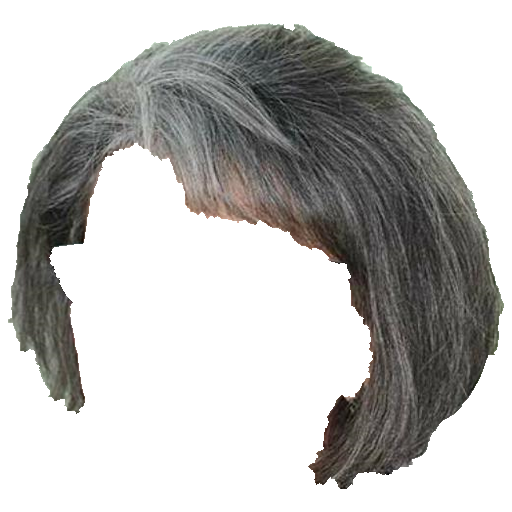}}
		\end{minipage}
	}
	\caption{Qualitative comparisons of our method against alternative approaches, for the task of Sketch2Hair. } 
\label{fig:comparisons}
\end{figure*}

\subsection{Dataset Construction}
We built our dataset in the following manner. First, we collected 640 high-resolution photographs with clear hair strands from internet, where the hair area is restricted up to $512\times512$. Then for each image, the hair region is manually segmented and the mask is saved. For Sketch2Hair, we distributed the 640 images to 5 paid users, who were requested to draw a set of lines in the hair area according to the growth directions of strands, generating 640 sketch-image pairs. For super-resolution, we only downsampled the $512^2$ resolution images to obtain the input data. We randomly split the whole data into training and testing sets with a ratio of $4:1$. A glance of the dataset is illustrated in \figurename~\ref{fig:dataset}.

\subsection{Comparisons on Sketch2Hair}
For the task of Sketch2Hair, we based our comparisons on pix2pix~\cite{isola2017image} , pix2pixHD~\cite{wang2018high}, CRN~\cite{chen2017photographic} and pix2pixStyle. For pix2pix, we just used the official code while replacing U-net256 with U-net512. CRN is a single feedforward network without adversarial structure. We firstly trained it in $256^2$ resolution and then fine-tuned it in $512^2$ resolution. Pix2pixHD is an improved model of pix2pix for large scale image synthesis, so we also used its official code on $512^2$ resolution. For Pix2pixStyle, we trained it on the structure of pix2pix with an additional style loss. For fair comparison, the 4 models were trained thoroughly with our dataset, until the convergence of their losses.

\figurename~\ref{fig:comparisons} shows the qualitative comparisons of our approach against the above 4 methods. As seen, the basic pix2pix framework produced blurry artifacts. Pix2pixHD improved the appearance by better generating the textured structure, while noise still exists in the results. CRN produced the most visual-pleasing outputs, but fine details were missing. Pix2pixStyle has similar performances as pix2pixHD while it lacks of enough high frequency information. By contrast, our approach not only synthesized better structure but also generated finer details, producing a silky and glowing appearance.

We also applied a no-reference image quality score i.e. Naturalness Image Quality Evaluator (NIQE) to quantitatively measure the quality of the results. A smaller score indicates better perceptual quality. The evaluated scores demonstrate our approach outperforms the 4 models above as seen in Table~\ref{tb:niqe}.
\begin{table}[t]
\centering

\resizebox{0.9\linewidth}{!}{
\begin{tabular}{c|c|c|c|c}
\hline
pix2pix & CRN    & pix2pixHD & pix2pixStyle & ours   \\ \hline
7.5943  & 9.7551 & 7.4186    & 7.4069       & \textbf{7.3827} \\ \hline
\end{tabular}} \vspace{0.5mm}

\resizebox{0.85\linewidth}{!}{
\begin{tabular}{c|c|c}
\hline
                        & 4$\times$ (SRGAN as $G_b$) & 8$\times$ (ESRGAN as $G_b$) \\ \hline
$I_c$           & 9.5084                  & 7.7891                   \\ \hline
$I_f$ (ours) & \textbf{7.5229}                  & \textbf{7.6702} \\ \hline
\end{tabular}} \vspace{1mm}
\caption{NIQE scores of Sketch2Hair (top) and super-resolution results (bottom). Lower value represents better quality.}\label{tb:niqe}
\end{table}
We further conducted a user study to perceptually evaluate the performances of the models stated above. Specifically, we randomly picked up 20 images from the total 128 testing images and paired them with the corresponding outcomes of the above 4 approaches, obtaining 80 pairs. Then we shuffled these pairs and asked volunteers to choose the one with finer strands in each pair. 65 volunteers were involved, 48 of whom are male, 94\% are from 18-25 years old, and 6\% are from 26-35. The results show that in the comparisons with CRN, pix2pix, pix2pixHD, and pix2pixStyle, our approach was voted by 98.23\%, 94.38\%, 86.62\%, 69.62\% of volunteers separately.

\subsection{Comparisons on Hair Super-Resolution}
\begin{figure}
	\centering
	\subfigure[$I_g$]{
		\begin{minipage}[t]{0.22\linewidth}
			{\includegraphics[width=2cm]{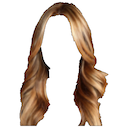}}
			{\includegraphics[width=2cm]{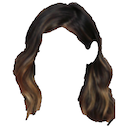}}
			{\includegraphics[width=2cm]{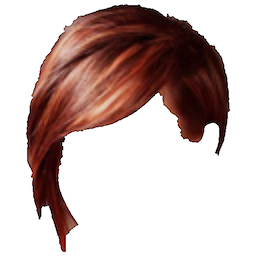}}
			{\includegraphics[width=2cm]{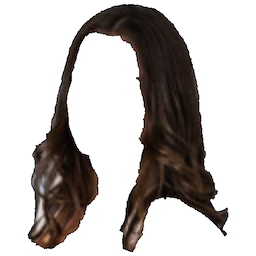}}
		\end{minipage}\label{fig:SRGAN:input}
	}
	\subfigure[$I_c$]{
		\begin{minipage}[t]{0.22\linewidth}
			{\includegraphics[width=2cm]{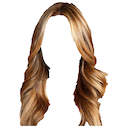}}
			{\includegraphics[width=2cm]{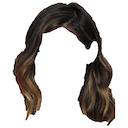}}
			{\includegraphics[width=2cm]{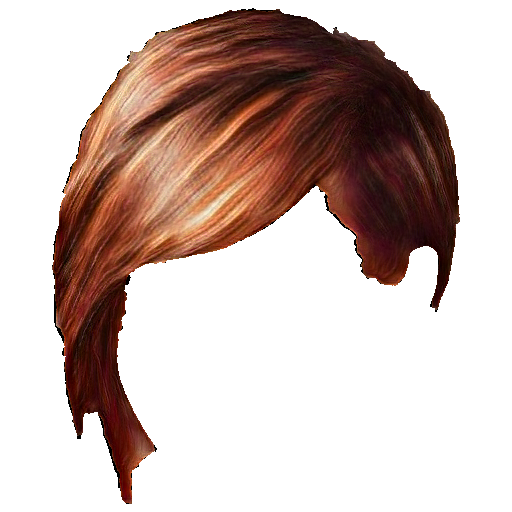}}
			{\includegraphics[width=2cm]{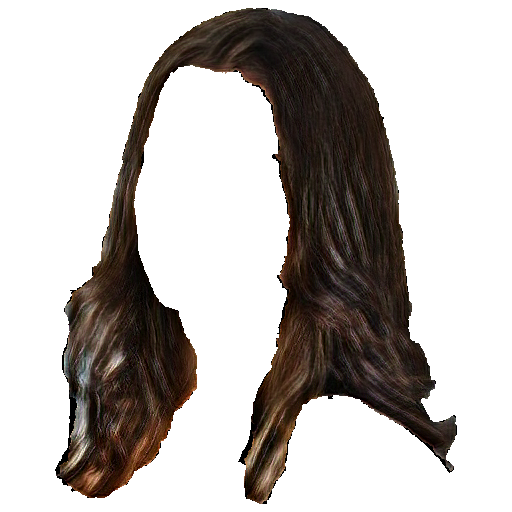}}
		\end{minipage}
	}
	\subfigure[$I_f$]{
		\begin{minipage}[t]{0.22\linewidth}
			{\includegraphics[width=2cm]{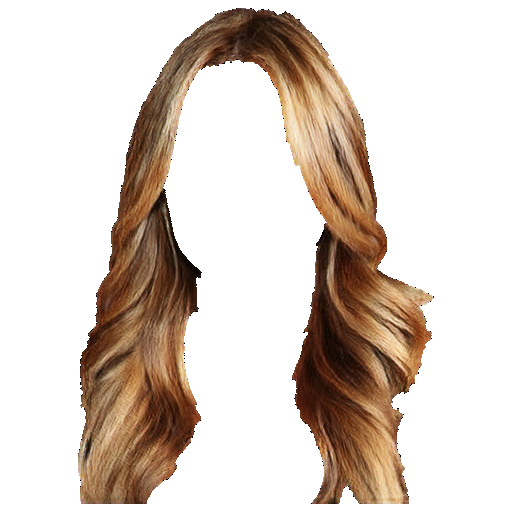}}
			{\includegraphics[width=2cm]{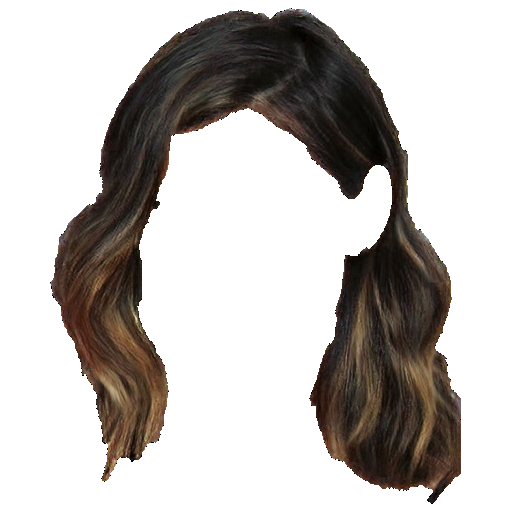}}
			{\includegraphics[width=2cm]{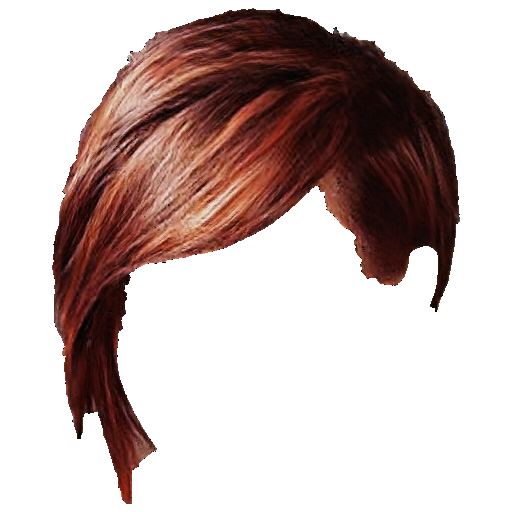}}
			{\includegraphics[width=2cm]{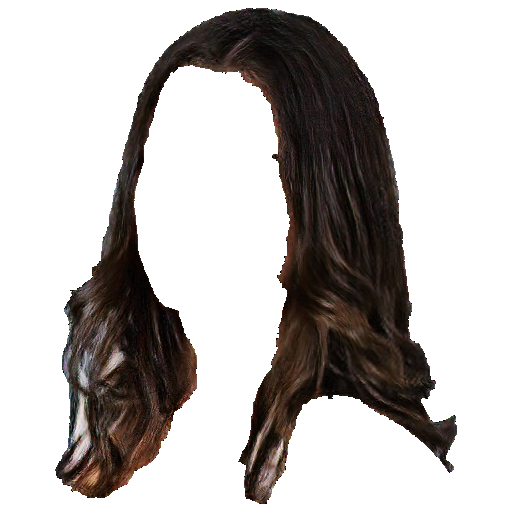}}
		\end{minipage}
	}
	\subfigure[$I_{gt}$]{
		\begin{minipage}[t]{0.22\linewidth}
			{\includegraphics[width=2cm]{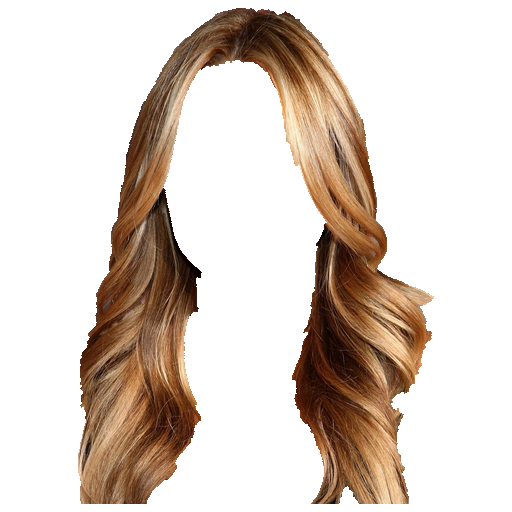}}
			{\includegraphics[width=2cm]{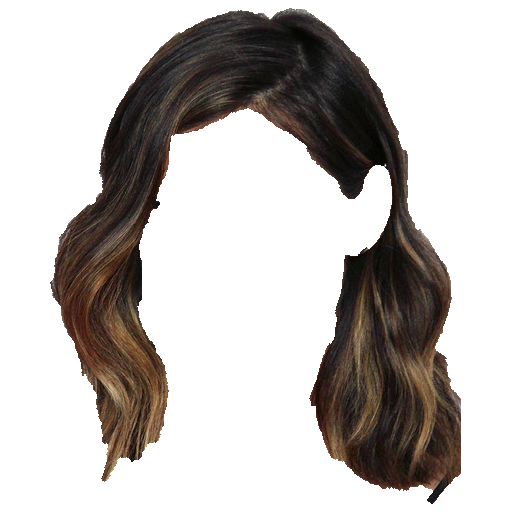}}
			{\includegraphics[width=2cm]{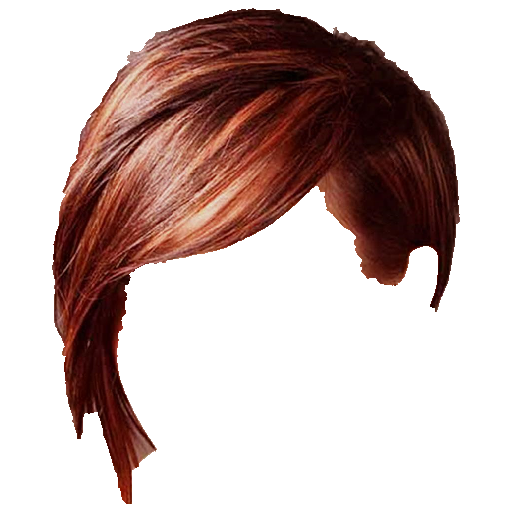}}
			{\includegraphics[width=2cm]{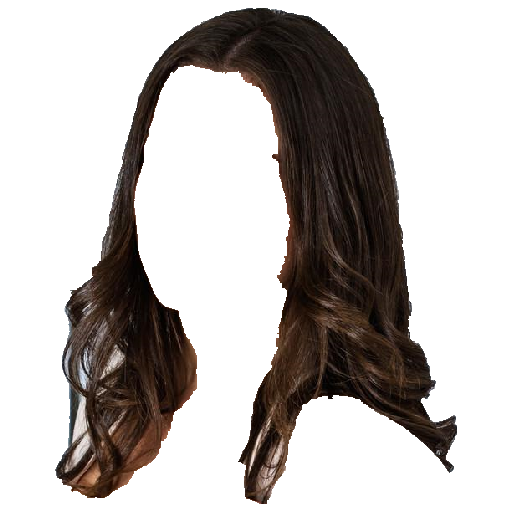}}
		\end{minipage}
	}
	\caption{Results by our approach and SRGAN / ESRGAN in the task of hair image super-resolution. (a) Low-resolution input. (b) SR results in 4x by SRGAN (top 2 rows) and in 8x by ESRGAN (bottom 2 rows). (c) Our enhanced results from (b). (d) Ground-truth. Enhanced hair are clearer and more visually pleasant (Better watching via zooming in). }
\label{fig:SRGAN}
\end{figure}

For the task of Hair Super-Resolution, we compared our approach with two state-of-the-art methods, SRGAN~\cite{Ledig2017Photo} and ESRGAN~\cite{Wang2018ESRGAN}, in 4x scale and 8x scale separately. We list the results in \figurename~\ref{fig:SRGAN}. As seen, SRGAN recovered limited image details in 4x results but the details are still fuzzy. ESRGAN produced clear enough but gummy hairlines in the 8x results. In contrast, our approach significantly reduced these artifacts and accurately reconstructed the finer details, producing plausible and visually pleasant results.

Similarly, NIQE was also applied here to quantitatively measure the quality of the results. We list the scores of results $I_c$ by SRGAN and ESRGAN, as well as the enhanced results $I_f$ by our method in Table.~\ref{tb:niqe}, which demonstrate the self-enhancing capability of our re-generating network.

\section{Conclusion and Discussion}\label{sec:conclusion}
We have presented a hair image synthesis approach given an image with limited guidance, such as sketch or low-resolution image. We generalize the hair synthesis in two phases. In Phase 1, we apply an existing image-to-image translation network to generate a coarse result, which is recognizable but lacks of textured details. Then we apply a re-generating network with self-enhancing capability to the coarse result, and produce the final high-quality  result. The self-enhancing capability is achieved by a proposed structure extraction layer, which extracts the texture and orientation map from a hair image. Experimental results demonstrated that our method outperforms state-of-the-art, in perceptual user study, qualitative and qualitative comparisons. We hope that this two-phase approach could be potentially applied to more hair-synthesis works.

Our results, while significantly outperforms the state-of-the-art in the realism, are still distinguishable from real photographs. Also, our method lacks the ability to control the shading and depth information of the generated hair, as no such information is available in the input sketch. Also, our method cannot handle complex hairstyles such as braids. Techniques like combining multi-modal features together as in~\cite{wang2014video,ren2016look} could be potential solutions. We will leave all of the goals as our future work.

\bibliographystyle{named}
\bibliography{ijcai19}

\end{document}